\documentclass[10pt,twocolumn,letterpaper]{article}

\usepackage{latex/cvpr}
\usepackage{times}
\usepackage{epsfig}
\usepackage{graphicx}
\usepackage{amsmath}
\usepackage{amssymb}

\usepackage{subfig}
\usepackage{booktabs}
\usepackage{multirow}
\usepackage{dblfloatfix}    
\usepackage{array}
\usepackage[font=small,labelfont=bf]{caption}

\newcommand{\drule}{\specialrule{0.2pt}{1pt}{1pt}%
            \specialrule{0.2pt}{0pt}{\belowrulesep}%
            }
\newenvironment{myindentpar}[1]%
  {\begin{list}{}%
          {\setlength{\leftmargin}{#1}}%
          \item[]%
  }
  {\end{list}}
\newcommand*{\affaddr}[1]{#1} 
\newcommand*{\affmark}[1][*]{\textsuperscript{#1}}
\newcommand*{\email}[1]{\texttt{#1}}
\usepackage{kotex}

\usepackage[pagebackref=true,breaklinks=true,letterpaper=true,colorlinks,bookmarks=false]{hyperref}

\cvprfinalcopy 

\def\cvprPaperID{6529} 

\begin{document}
\title{Cars Can't Fly up in the Sky: Improving Urban-Scene Segmentation via\\Height-driven Attention Networks}

\author{
\textbf{Sungha Choi}\affmark[1,3]\quad\quad\textbf{Joanne T. Kim}\affmark[2,3]\quad\quad \textbf{Jaegul Choo}\affmark[4]
\vspace*{0.1cm}
\\
\affaddr{\affmark[1]A\&B Center, LG Electronics,\, Seoul, South Korea\\
\affmark[2]Lawrence Livermore National Laboratory,\, Livermore, CA, USA\\
\affmark[3]Korea University,\, Seoul, South Korea\quad\quad\affmark[4]KAIST,\, Daejeon, South Korea}\\
\small{\email{\{shachoi,tengyee\}@korea.ac.kr}\quad\email{jchoo@kaist.ac.kr}}\\
}

\maketitle
\thispagestyle{empty}

\begin{abstract}
\vspace*{-0.2cm}
  This paper exploits the intrinsic features of urban-scene images and proposes a general add-on module, called \emph{height-driven attention networks (HANet)}, for improving semantic segmentation for urban-scene images. It 
  emphasizes informative features or classes selectively according to the vertical position of a pixel. The pixel-wise class distributions are significantly different from each other among horizontally segmented sections in the urban-scene images. Likewise, urban-scene images have their own distinct characteristics, but most semantic segmentation networks do not reflect such unique attributes in the architecture. The proposed network architecture incorporates the capability exploiting the attributes to handle the urban-scene dataset effectively. We validate the consistent performance (mIoU) increase of various semantic segmentation models on two datasets when HANet is adopted.
  This extensive quantitative analysis demonstrates that adding our module to existing models is easy and 
  cost-effective.
  Our method achieves a new state-of-the-art performance on the Cityscapes benchmark with a large margin among ResNet-101 based segmentation models.
  Also, we show that the proposed model is coherent with the facts observed in the urban scene by visualizing and interpreting the attention map. 
  Our code and trained models are publicly available\footnote{\vspace*{-0.2cm}\url{https://github.com/shachoi/HANet}}.

\end{abstract}
\vspace{-0.5cm}
\begin{figure*}[ht]
  \centering\includegraphics[width=0.95\linewidth]{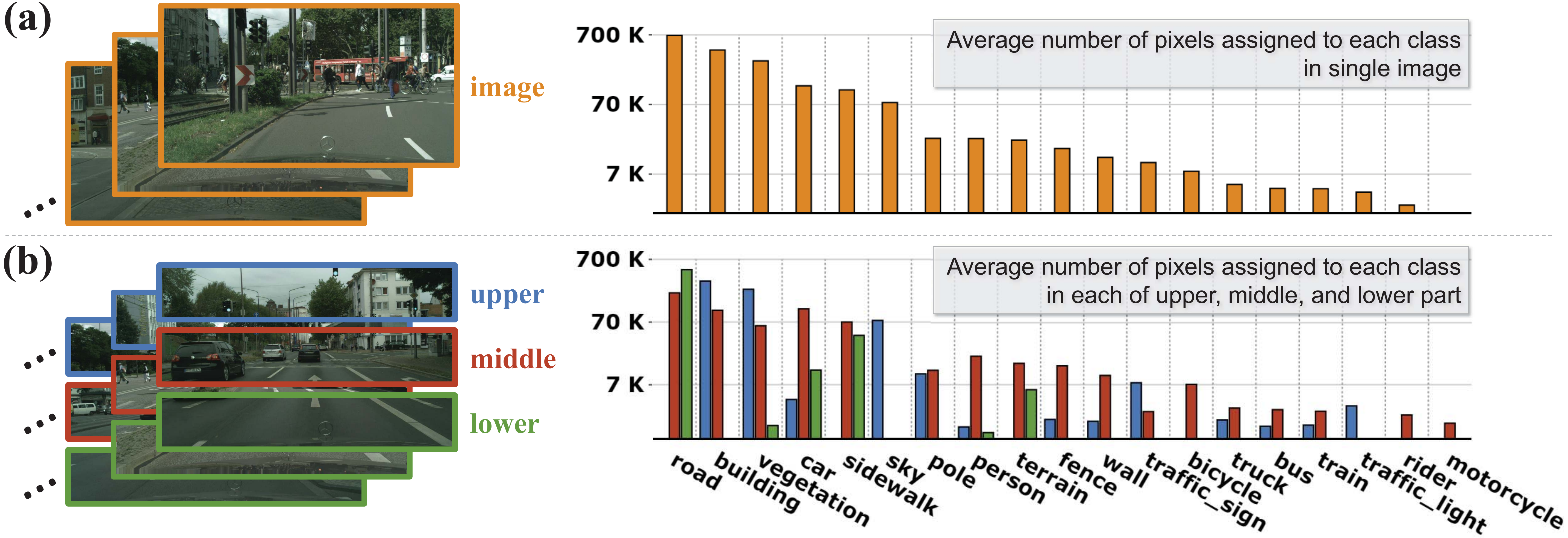}

 \vspace*{-0.4cm}
 \caption{Motivation of our approach, the pixel-wise class distributions. All numbers are average values obtained from the entire training set of the Cityscapes dataset with its pixel-level class labels~\cite{cordts2016cityscapes}. Note that there exists a total of 2048K pixels per image, and the y-axis is in log-scale.
 (a) Each bar
 represents the average number of pixels assigned to each class contained in a single image. 
For example, on average, about 685K pixels per image are assigned to the road class.
 (b) Each part of an image divided into three horizontal sections has a significantly different class distribution from each other. For example, the upper region has just 38 pixels of the road class, while the lower region has 480K pixels of it. }
 \label{fig:intro}
\vspace*{-0.7cm}
\end{figure*}

\section{Introduction} \label{sec:Intro}
\vspace{-0.1cm}
Semantic image segmentation, a fundamental task in computer vision, is employed for basic urban-scene understanding in autonomous driving. 
Fully convolutional networks (FCNs)~\cite{long2015fully} are 
seminal work that adopts deep convolutional neural networks (CNNs)
in semantic segmentation, by replacing fully connected layers with convolutional ones at the last stage in typical CNN architectures. Other advanced techniques, such as skip-connections in an encoder-decoder architecture~\cite{badrinarayanan2017segnet,ronneberger2015u,chen2018encoder}, an atrous convolution~\cite{chen2018deeplab, yu2015multi}, an atrous spatial pyramid pooling (ASPP)~\cite{chen2017rethinking}, and a pyramid pooling module~\cite{zhao2017pyramid}, have further improved the FCN-based architecture in terms of semantic segmentation performance. They have proven to be successful in diverse semantic segmentation benchmarks~\cite{everingham2010pascal,pascal-voc-2012,mottaghi2014role,cordts2016cityscapes,Alhaija2018IJCV,lin2014microsoft,neuhold2017mapillary} including urban-scene datasets.

Yet, urban-scene images have their own distinct nature related to perspective geometry~\cite{Li2017FoveaNetPU} and positional patterns~\cite{zou2018unsupervised,chen2018road}. Due to the fact that the urban-scene images are captured by the cameras mounted 
on the front side of a car, the urban-scene datasets consist only of road-driving pictures. This leads to the possibility of incorporating common structural priors depending on a spatial position, markedly 
in a vertical position. To verify this characteristic, we present the class distribution of an urban-scene dataset across vertical positions in Fig.~\ref{fig:intro}. Although the pixels of the few classes
are dominant in an entire region of an image (Fig.~\ref{fig:intro}(a)), the class distribution has significant dependency on a vertical position. That is, a lower part of an image is mainly composed of road, while 
the middle part contains various kinds of relatively small objects. 
In 
the upper part, buildings, vegetation, and sky are principal objects as shown in Fig.~\ref{fig:intro}(b).
Table~\ref{tab_intro} shows the probabilities with respect to the dominant top-5 classes: road, building, vegetation, car, and sidewalk. The class distribution is extremely imbalanced that the dominant classes take over 88\% of the entire dataset. As mentioned above, the class distribution is completely different if the image is divided into 
three regions: upper, middle, and lower parts. For instance, the probability of the road class $p_{\text{road}}$ is 36.9\% on average given an entire image, but this chance drops dramatically to 0.006\% for an upper region, while jumps to 87.9\% if a lower region is considered. 

Also, we analyzed this observation using entropy, a measure of uncertainty. The entropy of the probability distribution $X$ of a pixel over 19 classes in the Cityscapes dataset~\cite{cordts2016cityscapes} is computed as\\
\vspace*{-0.5cm}
\begin{align}
    H(X)&=H(p_{\text{road}}, p_{\text{building}}, \dots, p_{\text{motorcycle}}) \nonumber \\
    &=-\sum_{i}p_{i}\log{p_{i}},
\label{eq_entropy}
\end{align}
where $p_{i}$ denotes the probability that an arbitrary pixel is assigned to the $i$-th class.
the conditional entropy $H(X$\big|image$)$ of $X$ given an image, computed by Eq.~\eqref{eq_entropy}, is 1.84. On the other hand, the average conditional entropy of $X$ given each of the three regions as $H(X$\big|upper$)$, $H(X$\big|middle$)$, and $H(X$\big|lower$)$, is 1.26 as shown in Table~\ref{tab_intro}. As a result, one can see the uncertainty is reduced if we divide an image into several parts horizontally. Based on this analysis, if we can identify the part of an image to which a given arbitrary pixel belongs, it will be helpful for pixel-level classification in semantic segmentation.
\begin{table}[!b]
\vspace*{-0.4cm}
\begin{center}
\footnotesize
\begin{tabular}{c|ccccc|c|c}
\toprule
\multirow{2}{*}{Given} & \multicolumn{5}{c|}{Probabilities of top-5 classes} & \multicolumn{2}{c}{Entire class} \\\cline{2-6}
& $p_\text{road}$ & $p_\text{build.}$  & $p_\text{veg}$ & $p_\text{car}$ & $p_\text{swalk}$ & \multicolumn{2}{c}{entropy}\\
\drule
Image & 36.9 & 22.8 & 15.9 & 7.0 & 6.1 & \multicolumn{2}{c}{1.84} \\
\midrule
Upper & 0.006 & 47.8 & 35.4 & 0.6 & 0.009 & 1.24 & \multirow{3}{*}{\shortstack{1.26\\(avg)}}\\
Middle & 31.4 & 16.6 & 9.4 & 17.4 & 10.7 & 2.04 \\
Lower & 87.9 & 0.1 & 0.3 & 2.2 & 7.9 & 0.51 \\
\bottomrule
\end{tabular}
\end{center}
\vspace*{-0.5cm}
\caption{Comparison of the probability distributions (\%) of pixels being assigned to each class when 
an entire image is separated on upper, middle, and lower regions of the Cityscapes dataset.}
\label{tab_intro}
\end{table}

Motivated by these observations, we propose a novel height-driven attention networks
(HANet) as a general add-on module to semantic segmentation for urban-scene images. Given an input feature map, HANet extracts ``height-wise contextual information'', which represents the context of each horizontally divided part, and then predicts which features or classes are more important than the others within each horizontal part from the height-wise contextual information. 
The models adopting HANet consistently outperform baseline models. Importantly, our proposed module can be added to any CNN-based backbone networks with negligible cost increase. To verify the effectiveness of HANet, we conduct extensive experiments with various backbone networks such as ShuffleNetV2~\cite{ma2018shufflenet}, MobileNetV2~\cite{Sandler_2018}, ResNet-50~\cite{he2016deep}, and ResNet-101~\cite{he2016deep}. We also focus on lightweight backbone networks where a lightweight HANet is more effective.
The main contributions of this paper include:

\begin{myindentpar}{0.2cm}
\vspace*{-0.3cm}
\noindent$\bullet$ \,We propose a novel lightweight add-on module, called HANet,
which can be easily added to existing models and improves the performance by scaling the activation of channels according to the vertical position of a pixel. We show the effectiveness and wide applicability of our method through extensive experiments by applying on various backbone networks and two different datasets.
\vspace*{-0.1cm}
\\[0.3em]
$\bullet$ \,By adding HANet to the baseline, with negligible computational and memory overhead, we achieve a new state-of-the-art performance on the Cityscapes benchmark with a large margin among ResNet-101 based models.
\\[0.3em]
$\bullet$ \,We visualize and interpret the attention weights on individual channels
to experimentally confirm our intuition and rationale that height position is crucial to improve the segmentation performance on urban scene.
\end{myindentpar}

\vspace*{-0.4cm}
\section{Related Work}
\vspace*{-0.1cm}
\paragraph{Model architectures for semantic segmentation.}
Maintaining the resolution of a feature map while capturing high-level semantic features is essential in achieving high performance of semantic segmentation. Typically, high-level features are extracted by stacking multiple convolutions and spatial pooling layers, but the resolution gets coarser in the process. Several studies~\cite{long2015fully,noh2015learning} address this limitation by 
leveraging deconvolution for learnable upsampling from low-resolution features. Skip-connections overcome the limitation by recovering the object boundaries in a decoder layer through leveraging high-resolution features existing earlier in the encoder layer~\cite{badrinarayanan2017segnet,ronneberger2015u,lin2017refinenet,chen2018encoder}. Another prevalent method is atrous convolution~\cite{chen2018deeplab, yu2015multi}, which increases the receptive field size without increasing the number of parameters, and it is widely adopted in recent semantic segmentation networks~\cite{chen2017rethinking, chen2018encoder, yang2018denseaspp, zhang2018context, zhu2019improving}. Additionally, ASPP~\cite{chen2017rethinking} and pyramid pooling modules~\cite{zhao2017pyramid} address such challenges caused by diverse scales of objects.

More recently, long-range dependency is captured~\cite{zhu2019asymmetric,huang2019ccnet,fu2019dual} to improve the performance especially by extending self-attention mechanism~\cite{vaswani2017attention}.
Focusing on boundary information is another approach in semantic segmentation~\cite{bertasius2015high,chen2016semantic,ding2019boundary,takikawa2019gated,marin2019efficient}. Recent work imposes separate modules for targeting boundary processing~\cite{ding2019boundary,takikawa2019gated} or boundary driven adaptive downsampling~\cite{marin2019efficient}.
Also, capturing contextual information is widely exploited. ACFNet~\cite{zhang2019acfnet} uses class center to gather features of pixels in each class as categorical context, while CFNet~\cite{zhang2019co} learns the distribution of co-occurrent features and captures co-occurrent context to relook before making predictions. CiSS-Net~\cite{zhou2019context} adopts reinforcement learning to explore the context information in predicted segmentation maps, not having any supervision.


\vspace*{-0.1cm}
\paragraph{Exploitation of urban-scene image.}
\vspace*{-0.4cm}
In the field of semantic segmentation, several studies exploit the characteristics of the urban-scene images. In general, the scale of objects significantly vary in the urban-scene images. FoveaNet~\cite{Li2017FoveaNetPU} localizes a ``fovea region'', where the small-scale objects are crowded, and performs scale normalization to address heterogeneous object scales. DenseASPP~\cite{yang2018denseaspp} adopts densely connected ASPP, which connects multiple atrous convolutional layers~\cite{huang2017densely} to address large-scale changes of the objects. Another recent approach~\cite{zhu2019improving} exploits the fact that the urban-scene images have continuous video frame sequences and proposes the data augmentation technique based on a video prediction model to create future frames and their labels.


Recent approaches in the field of domain adaptation propose the method to leverage the properties of the urban-scene images. A class-balanced self-training with spatial priors~\cite{zou2018unsupervised} generates pseudo-labels for unlabeled target data by exploiting which classes appear frequently at a particular position in an image for unsupervised domain adaptation. Another approach~\cite{chen2018road} divides an urban-scene image into several spatial regions and conducts domain adaptation on the pixel-level features from the same spatial region. 
Also, 
the correlation between depth information and semantic is exploited to gain additional information from synthetic data for urban-scene domain adaptation~\cite{chen2019learning}.

\paragraph{Channel-wise attention.}
\vspace*{-0.5cm}
Our proposed method, HANet, has strong connections to a channel-wise attention approach, which exploits the inter-channel relationship of features and scales the feature map according to the importance of each channel. Squeeze-and-excitation networks (SENets)~\cite{hu2018squeeze} capture 
the global context of the entire image using global average pooling and predict per-channel scaling factors to extract informative features for an image classification task. This mechanism is widely adopted in subsequent studies~\cite{woo2018cbam,zhang2018context,yu2018learning,li2018pyramid} for image classification and semantic segmentation tasks. Inspired from ParseNet~\cite{liu2015parsenet}, which shows the impact of a global context of an entire image
in semantic segmentation, previous work~\cite{zhang2018context,yu2018learning,li2018pyramid} for semantic segmentation exploits the global context of the entire image to generate channel-wise attention. However, the urban-scene datasets~\cite{cordts2016cityscapes,Alhaija2018IJCV,neuhold2017mapillary,BrostowSFC:ECCV08} consist only of road-driving pictures, which means that the images share similar class statistics. Therefore, the global context should be relatively similar among urban-scene images. As a result, the global context of the entire image cannot present distinct information of each image to help per-pixel classification in urban-scene images. This explains why the previous work related to channel-wise attention for semantic segmentation mainly focuses on the generic scene datasets.

\begin{figure*}[ht!]
\vspace*{-0.2cm}
  \centering\includegraphics[width=0.97\linewidth]{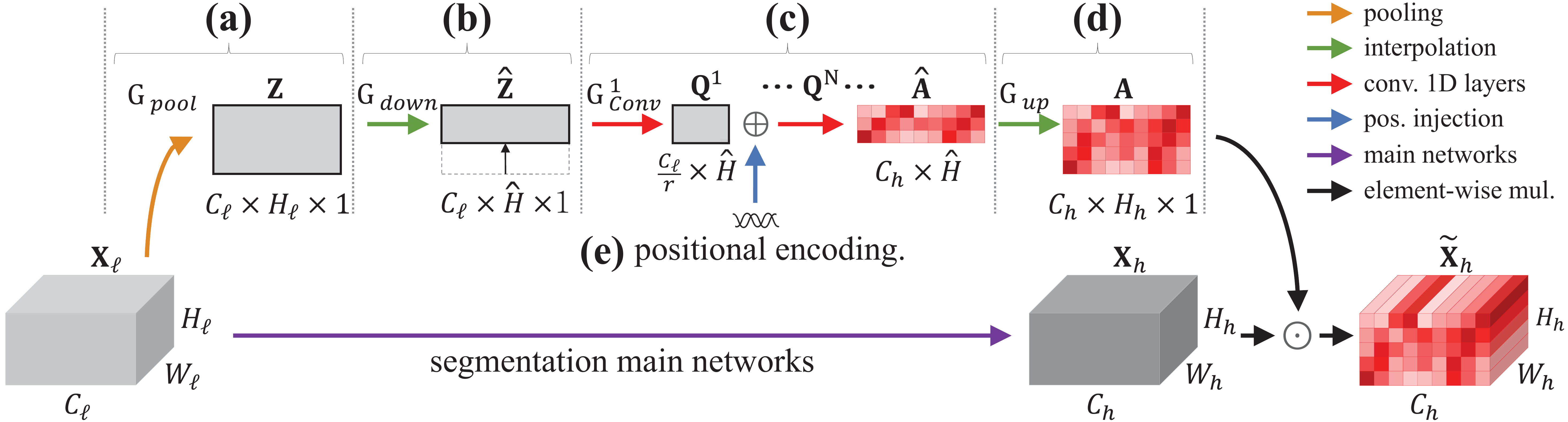}
\vspace*{-0.3cm}
  \caption{Architecture of our proposed HANet. 
  Each operation $op$ is notated as $\text{G}_{op}$, and feature maps are in bold--$\text{X}_{\ell}$: lower-level feature map, Z: width-wise pooled $\text{X}_{\ell}$, $\hat{\text{Z}}$: down-sampled Z, $\text{Q}^{n}$: $n$-th intermediate feature map of 1D convolution layers, $\hat{\text{A}}$: down-sampled attention map, A: final attention map, $\text{X}_{h}$: higher-level feature map, $\Tilde{\text{X}}_{h}$: transformed new feature map. Details can be found in Section~\ref{sec:method_block}.}
\label{fig:roab_arch}
\vspace*{-0.5cm}
\end{figure*}

\vspace*{-0.1cm}
\section{Proposed Method}
\vspace*{-0.1cm}
Urban-scene images generally involve common structural priors depending on a spatial position. Each row of an image has significantly different statistics in terms of a category distribution. In this sense, individually capturing the height-wise contextual information, which represents the global context of each row can be used to estimate how channels should be weighted during pixel-level classification for urban-scene segmentation. Therefore, we propose HANet
which aims to i) extract the height-wise contextual information and ii) compute height-driven attention weights to represent the importance of features (at intermediate layers) or classes (at last layer) for each row using the context.
In this section, we first describe HANet as 
a general add-on module and then present the semantic segmentation networks incorporating several HANet at different layers specialized for urban-scene segmentation.

\vspace*{-0.1cm}
\subsection{Height-driven Attention Networks (HANet)}\label{sec:method_block}
\vspace*{-0.1cm}
HANet generates per-channel scaling factors for each individual row from its height-wise contextual information as 
its architecture illustrated in Fig.~\ref{fig:roab_arch}.

Let $\mathbf{X}_\ell\in\mathbb{R}^{C_\ell\times H_\ell\times W_\ell}$ and $\mathbf{X}_h\in\mathbb{R}^{C_h\times H_h\times W_h}$ denote the lower- and higher-level feature maps in semantic segmentation networks, where $C$ is the number of channels, $H$ and $W$ are the spatial dimensions of the input tensor, height and width, respectively. Given the lower-level feature map $\mathbf{X}_\ell$, $\text{F}_{\text{HANet}}$ generates a channel-wise attention map $\mathbf{A}\in\mathbb{R}^{C_h\times H_h}$ made up of height-wise per-channel scaling factors and fitted to the channel and height dimensions of the higher-level feature map $\mathbf{X}_h$. This is done in a series of steps: width-wise pooling (Fig.~\ref{fig:roab_arch}(a)), interpolation for coarse attention (Fig.~\ref{fig:roab_arch}(b,d)), and computing height-driven attention map (Fig.~\ref{fig:roab_arch}(c)). Moreover, adding positional encoding is included in the process (Fig.~\ref{fig:roab_arch}(e)).

After computing the attention map, the given higher-level feature map $\mathbf{X}_h$ can be transformed into a new representation $\mathbf{\Tilde{X}}_h$, acquired by an element-wise multiplication of $\mathbf{A}$ and $\mathbf{X}_h$. Note that single per-channel scaling vector is derived for each individual row or for each set of multiple consecutive rows, so the vector is copied along with the horizontal direction, which is formulated as 
\begin{equation}
    \mathbf{\Tilde{X}}_h=\text{F}_{\text{HANet}}\left(\mathbf{X}_{\ell}\right)\odot\mathbf{X}_h=\mathbf{A}\odot\mathbf{X}_h.
\end{equation}    

\paragraph{Width-wise pooling (Fig.~\ref{fig:roab_arch}(a)).} \label{pooling}
\vspace*{-0.2cm}
In order to obtain a channel-wise attention map, we firstly extract height-wise contextual information from each row by aggregating the $C_\ell\times H_\ell\times W_\ell$ input representation $\mathbf{X}_\ell$ into a $C_\ell\times H_\ell\times 1$ matrix $\mathbf{Z}$ using a width-wise pooling operation $\mathbf{\text{G}_{\text{pool}}}$, i.e., 
\vspace*{-0.1cm}
\begin{equation}
    \mathbf{Z}=\text{G}_{\text{pool}}\left(\mathbf{X}_\ell\right).
\vspace*{-0.1cm}
\end{equation}
There are two typical pooling methods, average pooling and max pooling, to squeeze the spatial dimension. The choice between max pooling and average pooling for the width-wise pooling operation is a hyper-parameter and is empirically set to average pooling. Formally, the $h$-th row vector of $\mathbf{Z}$ is computed as
\begin{equation}
    \mathbf{Z}_{:,h}=[\frac{1}{W}\sum_{i=1}^W\mathbf{X}_{1,h,i};\dots;\frac{1}{W}\sum_{i=1}^W\mathbf{X}_{C,h,i}].
\end{equation}

\paragraph{Interpolation for coarse attention (Fig.~\ref{fig:roab_arch}(b,d)).} \label{coarse}
\vspace*{-0.2cm}
After the pooling operation, the model generates a matrix $\mathbf{Z}\in\mathbb{R}^{C_\ell\times H_\ell}$. However, not all the rows of matrix $\mathbf{Z}$ may be necessary for computing an effective attention map. As illustrated in Fig.~\ref{fig:intro}(b), class distributions for each part highly differ from each other, even if we divide the entire area into just three parts. Therefore, we interpolate $C_\ell\times H_\ell$ matrix $\mathbf{Z}$ into $C_\ell\times \hat{H}$ matrix $\mathbf{\hat{Z}}$ by downsampling it (Fig.~\ref{fig:roab_arch}(b)).
$\hat{H}$ is a hyper-parameter and is empirically set to 16. Since the attention map, constructed from downsampled representations, is also coarse, the attention map is converted to have the equivalent height dimension with the given higher-level feature map $\mathbf{X}_h$ via upsampling (Fig.~\ref{fig:roab_arch}(d)).
\vspace*{-0.1cm}
\paragraph{Computation of height-driven attention map (Fig.~\ref{fig:roab_arch}(c)).} \label{reduction}
\vspace*{-0.3cm}
A height-driven channel-wise attention map $\mathbf{A}$ is obtained by convolutional layers that take width-wise pooled and interpolated feature map $\mathbf{\hat{Z}}$ as input. Recent work that utilized a channel-wise attention in classification and semantic segmentation~\cite{hu2018squeeze, woo2018cbam, zhang2018context} adopts fully connected layers rather than convolutional layers since they generate a channel-wise attention for an entire image. However, we adopt convolutional layers to let the relationship between adjacent rows be considered while estimating the attention map since each row is related 
to its adjacent rows.

The attention map $\mathbf{A}$ indicates which channels are critical at each individual row. There may exist multiple informative features at each row in the intermediate layer; in the last layer, each row can be associated with multiple labels (\eg, road, car, sidewalk, etc.). To allow these multiple features and labels, a sigmoid function is used in computing the attention map, not a softmax function. These operations consisting of $N$ convolutional layers can be written as 
\begin{equation}
    \mathbf{A}=\text{G}_{\text{up}}\left(\sigma\left(\text{G}_{\text{Conv}}^N\left(\dotsm\delta\left(\text{G}_{\text{Conv}}^1\big(\mathbf{\hat{Z}}\big)\right)\right)\right)\right),
\end{equation}
where $\sigma$ is a sigmoid function, $\delta$ is a ReLU activation, and $\text{G}_{\text{Conv}}^i$ denotes $i$-th one-dimensional convolutional layer. We empirically adopt three convolutional layers:
the first one $\text{G}_{\text{Conv}}^{1}\big(\mathbf{\hat{Z}}\big)=\mathbf{Q}^1\in\mathbb{R}^{\frac{C_\ell}{r}\times \hat{H}}$ for channel reduction, 
the second one $\text{G}_{\text{Conv}}^{2}\left(\delta(\mathbf{Q}^1)\right)=\mathbf{Q}^2\in\mathbb{R}^{2\cdot\frac{C_\ell}{r}\times \hat{H}}$,
and the last one $\text{G}_{\text{Conv}}^{3}\left(\delta(\mathbf{Q}^2)\right)=\mathbf{\hat{A}}\in\mathbb{R}^{C_h\times \hat{H}}$ for generating an attention map.
The reduction ratio $r$ reduces the parameter overhead of HANet as well as gives a potential regularization effect. An analysis on the effect of various reduction ratio as a hyper-parameter will be presented in Section~\ref{exp_abl}. 

\vspace*{-0.1cm}
\paragraph{Incorporating positional encoding (Fig.~\ref{fig:roab_arch}(e)).}
\vspace*{-0.3cm}
When humans recognize a driving scene, they have 
prior knowledge on the vertical position of particular objects
(e.g., road and sky appear in the lower and upper part, respectively). Inspired by this observation, we add the sinusoidal positional encodings~\cite{vaswani2017attention} to the intermediate feature map $\mathbf{Q}^i$ at the $i$-th layer in the HANet. A hyper-parameter $i$ is analyzed in the supplementary material.
For injecting positional encodings, we follow the strategy proposed in Transformer~\cite{vaswani2017attention}. The dimension of the positional encodings is same as the channel dimension $C$ of the intermediate feature map $\mathbf{Q}^i$. The positional encodings are defined as
\vspace*{-0.1cm}
\begin{align*}
    PE_{(p, 2i)}&=sin\big(p/100^{2i/C}\big)\\
    PE_{(p, 2i+1)}&=cos\big(p/100^{2i/C}\big),
\end{align*}
where $p$ denotes the vertical position index in the entire image ranging from zero to $\hat{H}-1$ of coarse attention, and $i$ is the dimension. 
The number of 
the vertical position is set to $\hat{H}$ as the number of rows in coarse attention. The new representation $\Tilde{Q}$ incorporating positional encodings is formulated as
\begin{equation}
    \Tilde{Q}=Q\oplus PE,
\end{equation}
where $\oplus$ is 
an element-wise sum.

Height positions are randomly jittered by up to two positions to generalize over different camera location from various datasets to prevent an inordinately tight position-object coupling. Additionally, we experiment with using learnable positional embeddings~\cite{gehring2017convolutional} to find the best way to incorporate positional information in the supplementary material.

Meanwhile, CoordConv~\cite{liu2018intriguing} proposed to embed height and width coordinates in the intermediate features for various vision tasks: extra channels containing hard-coded coordinates (e.g., height, width, and optional $r$) are concatenated channel-wise to the input representation, and then a standard convolutional layer is applied. Unlike this model, HANet exploits positional information of height to obtain attention values, $C\times H \times 1$, which is used for gating the output representations of main networks. Therefore, HANet differs significantly from CoordConv in terms of how to exploit the positional information. We experimentally compare ours with CoordConv in Section~\ref{exp_abl}.

\vspace*{-0.1cm}
\subsection{Segmentation Networks based on HANet}
\begin{figure}[t]
\begin{center}
  \includegraphics[width=0.88\linewidth]{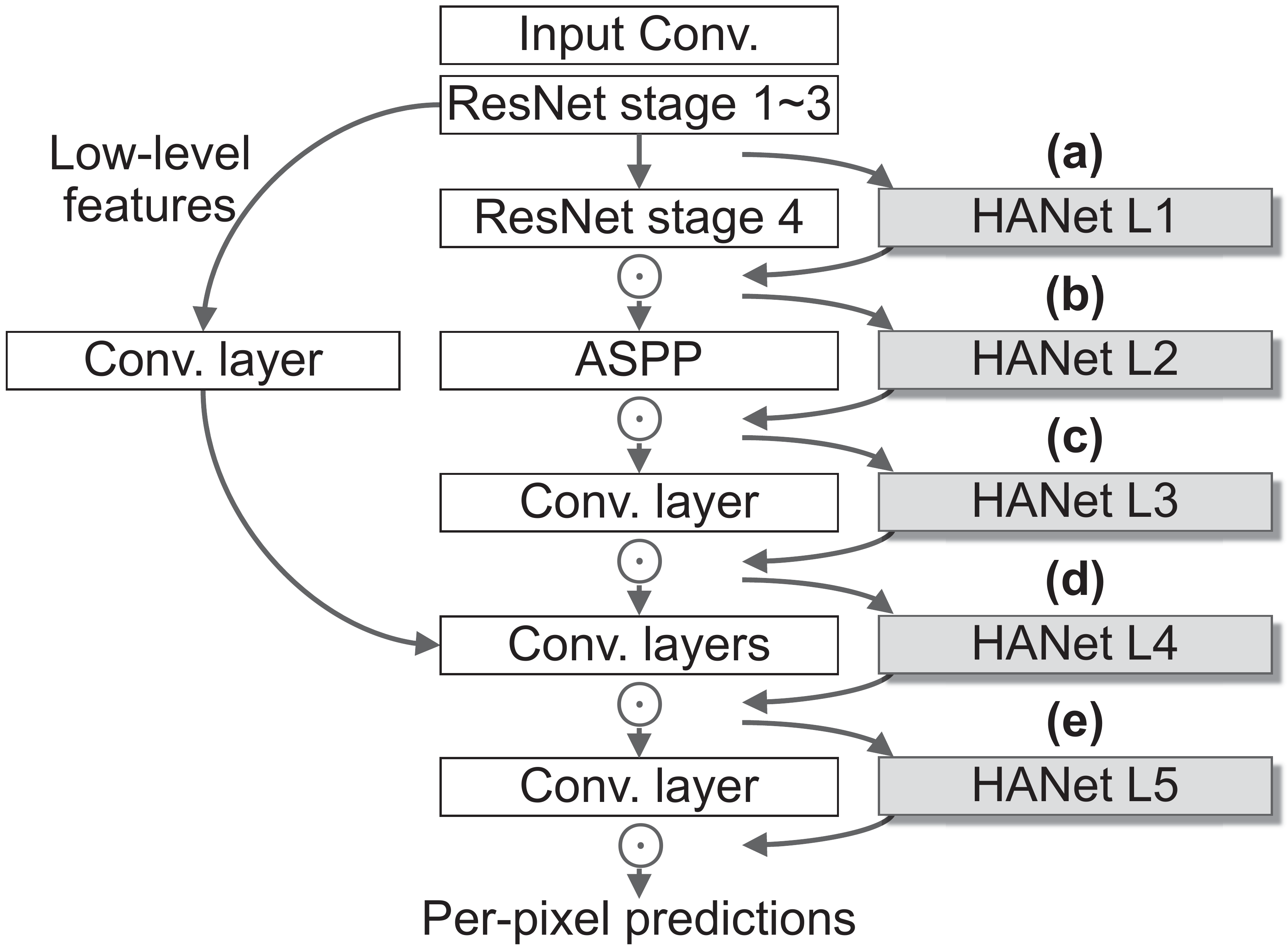}
\end{center}
\vspace*{-0.7cm}
   \caption{Semantic segmentation networks incorporating HANet in five different layers.}
\label{fig:sem_arch}
\vspace*{-0.6cm}
\end{figure}

\vspace*{-0.1cm}
We adopt DeepLabv3+~\cite{chen2018deeplab} as a baseline for semantic segmentation. DeepLabv3+ has an encoder-decoder architecture with ASPP that employs various dilation rates. We add HANet to the segmentation networks at five different layers (Fig.~\ref{fig:sem_arch}) after the point where high-level representation is encoded from backbone networks. 
This is because the higher-level feature has 
a stronger correlation with 
the vertical position. We conduct 
an ablation study to see the performance gain from adding HANet at different layers.
\vspace*{-0.1cm}
\subsection{Comparison with Other Attention Strategies}
\vspace*{-0.1cm}
Self-attention-based approaches like DANet~\cite{fu2019dual}, obtain attention values, ($H\times W$)$\times$($H\times W$), $C\times C$, from the semantic interdependencies in spatial and channel dimensions, respectively.
However, HANet does not consider the interdependencies among the dimensions. HANet includes one-dimensional convolutional layers being a separate branch as a modular design (Fig.~\ref{fig:roab_arch}(a)-(d)),
which specifically consider the structural property in urban-scene data. In this manner, HANet derives attention values, $C\times H\times 1$, to gate activation at the horizontal section of a feature map output in the main networks, considering vertical position. 
HANet is significantly more lightweight than self-attention-based approaches that considers the relationship between every pair in each dimension.
Meanwhile, channel-wise attention approaches such as SENet~\cite{hu2018squeeze} generates attention values, $C\times 1\times 1$, only at an image level. This is not ideal for urban-scene segmentation as most of the images shares similar circumstances and context.

\vspace*{-0.2cm}
\section{Experiments}\label{sec:Exp}
\vspace*{-0.1cm}

In this section, we first describe the implementation details of HANet. Then, we experimentally demonstrate 
the effectiveness and wide applicability of our proposed methods by extensive quantitative analysis including ablation studies. We evaluate HANet on two different urban-scene datasets including Cityscapes~\cite{cordts2016cityscapes} and BDD100K~\cite{yu2018bdd100k}. Furthermore, we visualize and analyze the attention map generated from HANet. For all the quantitative experiments, we measure the segmentation performance in terms of mean Intersection over Union (mIoU) metric.
\vspace*{-0.2cm}
\subsection{Experimental Setup}
\vspace*{-0.1cm}
\paragraph{Base segmentation architecture}
Our network architecture for semantic segmentation is based on DeepLabv3+~\cite{chen2018encoder}. We adopt various backbone networks including ShuffleNetV2~\cite{ma2018shufflenet}, MobileNetV2~\cite{Sandler_2018}, and ResNet~\cite{he2016deep} to verify wide applicability of HANet. Note that HANet can be easily inserted on top of various backbone networks. The adopted backbone networks are pretrained on ImageNet~\cite{russakovsky2015imagenet} for all the experiments, unless otherwise noted.


\vspace*{-0.5cm}
\paragraph{Stronger baseline}
To strictly verify the effectiveness of HANet, we improved the performance of the DeepLabv3+ baseline adopting ResNet-101, by integrating SyncBatchNorm (batch statistics synchronized across multiple GPUs) publicly included in PyTorch v1.1 and by replacing a single 7$\times$7 convolution by three 3$\times$3 convolutions in the first layer of ResNet-101. we also adopted an auxiliary cross-entropy loss in the intermediate feature map and class uniform sampling~\cite{rota2018place, zhu2019improving} to handle class imbalance problems. 
As a result, our baseline achieves the mIoU of 79.25\% on the Cityscapes validation set, which surpasses the other baseline models based on DeepLabv3+ architecture with ResNet-101 of previous work.

\vspace*{-0.5cm}
\paragraph{Training protocol.}
We employ SGD optimizer with initial learning rate of 1e-2 and momentum of 0.9. The weight decays are 5e-4 and 1e-4 for main networks and HANet, respectively. The learning rate scheduling follows the polynomial learning rate policy~\cite{liu2015parsenet}. The initial learning rate is multiplied by $\big(1-\frac{iteration}{max\_ iteration}\big)^{power}$, where power is 0.9. To avoid overfitting, typical data augmentations in semantic image segmentation models are used, including random horizontally flipping, random scaling in the range of [0.5,2], gaussian blur, color jittering, and random cropping.

\begin{table}[t]
\begin{center}
\setlength\tabcolsep{4.5pt}
\footnotesize
\begin{tabular}{c|c|c|ccc}
\toprule
Backbone & OS & Models & Params & GFLOPs & mIoU(\%) \\
\drule
\multirow{4}{*}{\shortstack{ShuffleNet\\V2 (1$\times$)~\cite{ma2018shufflenet}}} &  \multirow{2}{*}{32} & Baseline & 12.6M  & 64.34 & 70.27 \\
& & +HANet & 14.9M & 64.39 & \textbf{71.30} \\\cmidrule{2-6}
&  \multirow{2}{*}{16} & Baseline & 12.6M  & 117.09 & 70.85 \\
& & +HANet & 13.7M & 117.14 & \textbf{71.52} \\
\midrule
\multirow{4}{*}{\shortstack{MobileNet\\V2~\cite{ma2018shufflenet}}} &  \multirow{2}{*}{16} & Baseline & 14.8M  & 142.74 & 73.93 \\
& & +HANet & 16.1M & 142.80 & \textbf{74.96} \\\cmidrule{2-6}
&  \multirow{2}{*}{8} & Baseline & 14.8M  & 428.70 & 73.40 \\
& & +HANet & 15.4M & 428.82 & \textbf{74.70} \\
\midrule
\multirow{4}{*}{\shortstack{ResNet-50\\~\cite{he2016deep}}} & \multirow{2}{*}{16} & Baseline & 45.1M & 553.74 & 76.84  \\
& & +HANet & 47.6M & 553.85 & \textbf{77.78} \\\cmidrule{2-6}
& \multirow{2}{*}{8} & Baseline & 45.1M & 1460.56 & 77.76  \\
& & +HANet & 46.3M & 1460.76 & \textbf{78.71} \\
\midrule
\multirow{4}{*}{\shortstack{ResNet-101\\~\cite{he2016deep}}} & \multirow{2}{*}{16} & Baseline & 64.2M & 765.53 & 77.80 \\
& & +HANet & 65.4M & 765.63 & \textbf{79.31} \\\cmidrule{2-6}
& \multirow{2}{*}{8} & Baseline & 64.2M & 2137.82 & 79.25 \\
& & +HANet & 65.4M & 2138.02 & \textbf{80.29} \\
\bottomrule
\end{tabular}
\end{center}
\vspace*{-0.6cm}
\caption{Comparison of mIoU, the number of model parameters and FLOPs between the baseline and HANet on Cityscapes validation set according to various backbone networks and output stride. Adding HANet to the baseline consistently increase the mIoU with minimal cost increase.}
\label{tab_strong}
\vspace*{-0.65cm}
\end{table}



\vspace*{-0.15cm}
\subsection{Cityscapes}
\vspace*{-0.1cm}
The Cityscapes dataset~\cite{cordts2016cityscapes} is a large-scale urban-scene dataset, holding high-quality pixel-level annotations of 5K images and 20K coarsely annotated images. Finely annotated images consist of 2,975 train images, 500 validation images, and 1,525 test images. The annotations of test images are withheld for benchmarks. The resolution of each image is 2048$\times$1024, and 19 semantic labels are defined.
In all the experiments on Cityscapes validation set, we train our models using finely annotated training set for 40K iterations with a total batch size of 8 and a crop size of 768$\times$768.



\begin{figure}[t]
\begin{center}
\hspace*{-0.3cm}
  \includegraphics[width=0.99\linewidth]{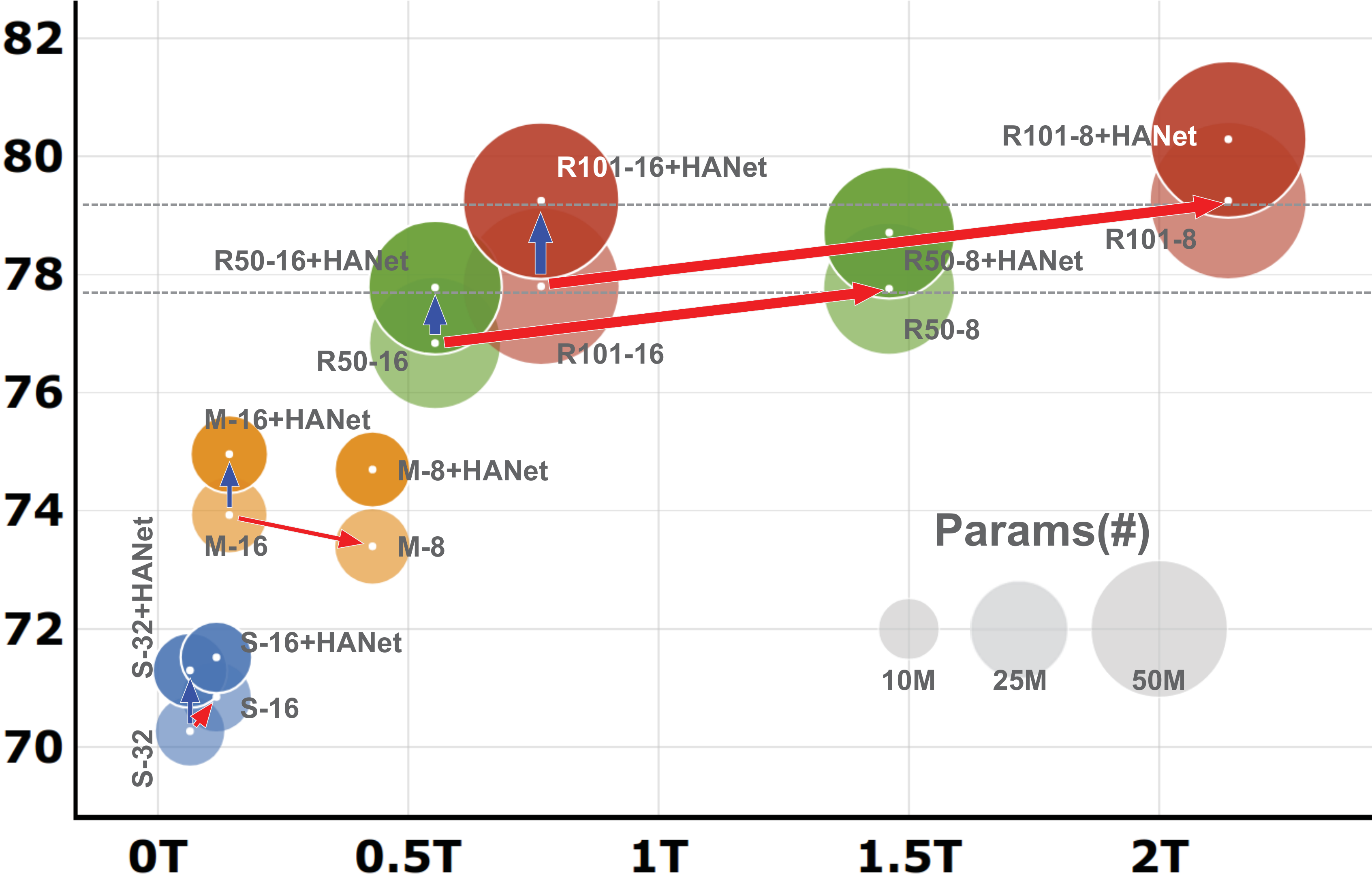}
\end{center}
\vspace*{-0.6cm}
   \caption{
   Comparison of the performance and complexity among the baseline and HANet on the various backbone networks.
   x-axis denotes teraFLOPs and y-axis denotes mIoU. The circle size denotes the number of model parameters. The texts in the colored circle indicate backbone networks, output stride, and whether HANet is adopted to the baseline. S, M, R50, and R101 denote ShuffleNetV2, MobileNetV2, ResNet-50, and -101, respectively. (e.g.,
   S-16: Baseline, ShuffleNetV2, and output stride 16)}
\label{fig:vis_scatter}
\vspace*{-0.5cm}
\end{figure}

\setcounter{table}{5}
\begin{table*}[!hb]
\vspace*{-0.3cm}
\begin{center}
\setlength\tabcolsep{4.0pt}
\footnotesize
\resizebox{\textwidth}{!}{
\begin{tabular}{c|c|c|cccccccccccccccccccc}
\toprule
Model (Year) & Backbone & mIoU & road & swalk & build. & wall & fence & pole & tligh. & tsign & veg & terr. & sky & pers. & rider & car & truck & bus & train & mcyc & bcyc\\
\drule
BFP~\cite{ding2019boundary}('19) & ResNet-101 & 81.4 & 98.7 & 87.0 & 93.5 & 59.8 & 63.4 & 68.9 & 76.8 & 80.9 & 93.7 & 72.8 & 95.5 & 87.0 & 72.1 & 96.0 & 77.6 & 89.0 & 86.9 & 69.2 & 77.6 \\
\midrule
DANet~\cite{fu2019dual}('19) & ResNet-101 & 81.5 & 98.6 & 86.1 & 93.5 & 56.1 & 63.3 & 69.7 & 77.3 & 81.3 & 93.9 & 72.9 & 95.7 & 87.3 & 72.9 & 96.2 & 76.8 & 89.4 & 86.5 & 72.2 & 78.2 \\
\midrule
CCNet~\cite{fu2019dual}('19) & ResNet-101 & 81.5 & 98.6 & 86.1 & 93.5 & 56.1 & 63.3 & 69.7 & 77.3 & 81.3 & 93.9 & 72.9 & 95.7 & 87.3 & 72.9 & 96.2 & 76.8 & 89.4 & 86.5 & 72.2 & 78.2 \\
\midrule
ACFNet~\cite{zhang2019acfnet}('19) & ResNet-101 & 81.8 & 98.7 & 87.1 & 93.9 & 60.2 & 63.9 & 71.1 & 78.6 & 81.5 & 94.0 & 72.9 & 95.9 & 88.1 & 74.1 & 96.5 & 76.6 & 89.3 & 81.5 & 72.1 & 79.2 \\
\midrule
Ours & ResNet-101 & \textbf{82.1} & 98.8 & 88.0 & 93.9 & 60.5 & 63.3 & 71.3 & 78.1 & 81.3 & 94.0 & 72.9 & 96.1 & 87.9 & 74.5 & 96.5 & 77.0 & 88.0 & 85.9 & 72.7 & 79.0 \\
\drule
DeepLabv3+~\cite{chen2018encoder}('18)$^\dagger$ & Xception~\cite{chollet2017xception} & 82.1 & 98.7 & 87.0 & 93.9 & 59.5 & 63.7 & 71.4 & 78.2 & 82.2 & 94.0 & 73.0 & 95.8 & 88.0 & 73.3 & 96.4 & 78.0 & 90.9 & 83.9 & 73.8 & 78.9 \\
\midrule
GSCNN~\cite{takikawa2019gated}('19)$^\ddagger$ & WideResNet38~\cite{zagoruyko2016wide} & 82.8 & 98.7 & 87.4 & 94.2 & 61.9 & 64.6 & 72.9 & 79.6 & 82.5 & 94.3 & 74.3 & 96.2 & 88.3 & 74.2 & 96.0 & 77.2 & 90.1 & 87.7 & 72.6 & 79.4 \\
\midrule
Ours$^\dagger$$^\ddagger$& ResNext-101~\cite{xie2017aggregated} & \textbf{83.2} & 98.8 & 88.0 & 94.2 & 66.6 & 64.8 & 72.0 & 78.2 & 81.4 & 94.2 & 74.5 & 96.1 & 88.1 & 75.6 & 96.5 & 80.3 & 93.2 & 86.6 & 72.5 & 78.7 \\
\bottomrule
\end{tabular}}
\end{center}
\vspace*{-0.5cm}
\caption{Comparison of mIoU and per-class IoU with other state-of-the-art models on Cityscapes test set. $^\dagger$ denotes training including Cityscapes coarsely annotated images. $^\ddagger$ denotes training with Mapillary pretrained model.
}
\label{tab_sota}
\end{table*}

\vspace*{-0.3cm}
\subsubsection{Effectiveness of the HANet components.}
\vspace*{-0.1cm}
Table~\ref{tab_strong} shows the effect of adopting HANet through performance increase (mIoU) according to the number of parameters and FLOPs which indicate model size and complexity, respectively.
To demonstrate the wide applicability of HANet, various backbones are examined including ShuffleNetV2~\cite{ma2018shufflenet}, MobileNetV2~\cite{Sandler_2018}, ResNet-50, and ResNet-101~\cite{he2016deep}.
Models with HANet consistently outperform baseline models with significant increases on MobileNetV2 and ResNet-101.
Moreover, the model parameters and complexity results indicate that the cost of adding HANet is practically negligible.
From Fig.~\ref{fig:vis_scatter}, we can see clearly that adding HANet (\textcolor{blue}{blue arrow}) is worth more than it costs in FLOPs, compared to improving the model through changing output stride (\textcolor{red}{red arrow}).
Therefore, HANet has a great advantage of not only an effective way of improving semantic segmentation accuracy, but also lightweight algorithm design for practical usage.


\vspace*{-0.3cm}
\subsubsection{Ablation studies} \label{exp_abl}
\vspace*{-0.2cm}
For ablation studies, we use ResNet-101 backbone with output stride of 16 and evaluate on the Cityscapes validation set.
Table~\ref{tab_ablation_component} shows the performance gain when HANet is added to the multiple layers and incorporates the positional encodings. Additionally, we conduct experiments by changing channel reduction ratio $r$ and pooling method. 
When we add HANet including positional encodings at multiple layers from L1 to L4 (Fig.~\ref{fig:sem_arch}) and the reduction ratio is set to 32, the mIoU significantly increases from 77.80 to 79.31.
To compare with CoordConv~\cite{liu2018intriguing}, we conduct experiments by replacing standard convolutional layers after the backbone with CoordConv. Table~\ref{tab_ablation_component} shows that HANet outperforms the baseline with CoordConv.

\setcounter{table}{2}
\begin{table}[!h]
\vspace*{-0.2cm}
\begin{center}
\footnotesize
\setlength\tabcolsep{5.0pt} 
\begin{tabular}{ccccc|c|c|c|c}
\toprule
\multicolumn{5}{c|}{Layers} & \multirow{2}{*}{\shortstack{Positional\\encoding}} & \multirow{2}{*}{Ratio $r$} & \multirow{2}{*}{\shortstack{Pooling\\method}} &\multirow{2}{*}{mIoU} \\
1 & 2 & 3 & 4 & 5 & & & \\
\drule
\checkmark & & & & & \checkmark & 32 & average & 78.52 \\
\checkmark & \checkmark & & & & \checkmark & 32 & average & 78.85 \\
\checkmark & \checkmark & \checkmark & & & \checkmark & 32 & average & 78.72 \\
\checkmark & \checkmark & \checkmark & \checkmark & & \checkmark & 32 & average & \textbf{79.31} \\
\checkmark & \checkmark & \checkmark & \checkmark & \checkmark & \checkmark & 32 & average & 78.79 \\
\midrule
\checkmark & \checkmark & \checkmark & \checkmark & & \checkmark & 16 & average & 79.15 \\
\checkmark & \checkmark & \checkmark & \checkmark & & \checkmark & 64 & average & 79.08 \\
\midrule
\checkmark & \checkmark & \checkmark & \checkmark & & & 32 & average & 78.25 \\
\midrule
\checkmark & \checkmark & \checkmark & \checkmark & & \checkmark & 32 & max & 78.87 \\
\midrule
\multicolumn{8}{c|}{Baseline} & 77.80 \\
\midrule
\multicolumn{8}{c|}{Baseline + CoordConv~\cite{liu2018intriguing} (Height + Width)} & 78.82 \\
\bottomrule
\end{tabular}
\end{center}
\vspace*{-0.5cm}
\caption{Ablation studies and hyper-parameter impacts with regard to the HANet injected layers, using positional encodings or not, and channel reduction ratio. ResNet-101, output stride 16 on Cityscapes validation set.}
\label{tab_ablation_component}
\vspace*{-0.6cm}
\end{table}

\vspace*{-0.1cm}
\subsubsection{Inference techniques.} \label{exp_inference}
\vspace*{-0.2cm}
For further performance improvements, we adopt frequently used techniques such as left-right flipped, multi-scale (with \textit{scales}=\{0.5, 1.0, 2.0\}) and sliding inference.
In such manner, our best model achieves 82.05\% mIoU on the Cityscapes validation set as in Table~\ref{tab_best_model}.

\begin{table}[h]
\vspace*{-0.2cm}
\begin{center}
\footnotesize
\begin{tabular}{c|c|c}
\toprule
Inference techniques & Baseline & +HANet \\
\drule
None & 79.25 & \textbf{80.29} \\ 
\midrule
Multiscale, Sliding, and Flipping & 81.14 & \textbf{82.05} \\
\bottomrule
\end{tabular}
\end{center}
\vspace*{-0.5cm}
\caption{mIoU(\%) comparison with respect to inference techniques. ResNet-101, output stride 8 on Cityscapes validation set.}
\label{tab_best_model}
\vspace*{-0.3cm}
\end{table}

\vspace*{-0.45cm}
\subsubsection{Efficacy at segmented regions}
\vspace*{-0.15cm}
As mentioned in Section~\ref{sec:Intro}, the average entropy decreases as we divide the image into multiple horizontal subsections. This implies the performance improvement in the entire region of an image. 
Besides, the entropy of the upper and lower regions have low entropy compared to the middle region. In this respect, we expect the performance increase arising from HANet would be larger in the upper and lower regions than that in the middle or entire region. Indeed, the performance significantly rises on the upper and lower regions as in Table~\ref{tab_ablation_parts}.

\begin{table}[h!]
\vspace*{-0.2cm}
\begin{center}
\footnotesize
\setlength\tabcolsep{0.5pt} 
\begin{tabular}{>{\centering}m{1.5cm}|>{\centering}m{1.25cm}>{\centering}m{1.25cm}>{\centering}m{1.25cm}>{\centering}m{1.25cm}|c}
\toprule
Model & Upper & Mid-upper & Mid-lower & Lower & \;\; Entire \;\\
\drule
Baseline & 78.69 & 76.35 & 83.16 & 70.59 & \;\; 81.14 \;\\ 
\midrule
+HANet & 80.29 & 77.09 & 84.09 & 73.04 & \;\; 82.05 \; \\
\drule
Increase(\%) & +\textbf{1.60} & +0.74 & +0.93 & +\textbf{2.45} & \;\; +0.91 \; \\
\bottomrule
\end{tabular}
\end{center}
\vspace*{-0.6cm}
\caption{mIoU(\%) comparison to baseline on each part of image divided into four horizontal sections. ResNet-101, output stride 8 on Cityscapes validation set.}
\label{tab_ablation_parts}
\vspace*{-0.6cm}
\end{table}


\subsubsection{Comparison to other state-of-the-art models}\label{sec:sota}
\vspace*{-0.2cm}
To compare with other state-of-the-arts, we train our models using finely annotated training and validation set for 90K iterations. In case of adopting ResNext-101~\cite{xie2017aggregated} backbone, coarsely annotated images are additionally used and the model is pretrained on Mapillary~\cite{neuhold2017mapillary}. The crop size and batch size are changed into 864$\times$864 and 12, respectively. The inference techniques from Section~\ref{exp_inference} are used, but we do not adopt any other techniques such as online bootstrapping for hard training pixels~\cite{wu2016highperformance}. We compare our best models based on ResNet-101 and ResNext-101 with other recent models on the Cityscapes test set (Table~\ref{tab_sota}). Our models achieve a new state-of-the-art performance.

\vspace*{-0.1cm}
\subsection{BDD100K}
\vspace*{-0.1cm}
Berkeley Deep Drive dataset (BDD100K)~\cite{yu2018bdd100k} is a large-scale diverse driving video database. It includes a semantic image segmentation dataset, consisting of 7,000 training and 1,000 validation images with a resolution of 1280$\times$720. 
It is a challenging dataset including images of various driving circumstance such as day, night, and diverse weather conditions.
Table \ref{tab_bdd100k} shows the superior results of HANet in BDD100K. 
The training strategy of BDD100K is similar to Cityscapes, but we change the crop size into 608$\times$608 and train for 60K iterations with a total batch size of 16. 
These adjustments simply comes from smaller image size and larger dataset size compared to Cityscapes.

\setcounter{table}{6}
\begin{table}[h!]
\begin{center}
\footnotesize
\begin{tabular}{c|c|c|c}
\toprule
Backbone & OS & Baseline & +HANet \\
\drule
MobileNetV2 & 16 & 58.91\% & \textbf{60.05\%} \\
\midrule
ResNet-101 & 16 & 63.75\% & \textbf{64.56\%} \\
\midrule
ResNet-101 & 8 & 64.84\% & \textbf{65.60\%} \\
\bottomrule
\end{tabular}
\end{center}
\vspace*{-0.5cm}
\caption{Comparison of mIoU between the baseline and HANet on BDD100K validation set. The output stride is set to 16 and the crop size is 608$\times$608.}
\label{tab_bdd100k}
\vspace{-0.4cm}
\end{table}




\vspace*{-0.1cm}
\subsection{Qualitative Analysis} \label{sec:vis}
\vspace*{-0.1cm}
\paragraph{Attention map visualization.}
We visualize the attention weights to analyze the behavior of the proposed HANet.  The attention visualization highlights those channels emphasized at a different vertical position.
Through the visualization, we can find out interesting clues to validate our observations and methods.

\begin{figure}[h]
\begin{center}
  \includegraphics[width=0.99\linewidth]{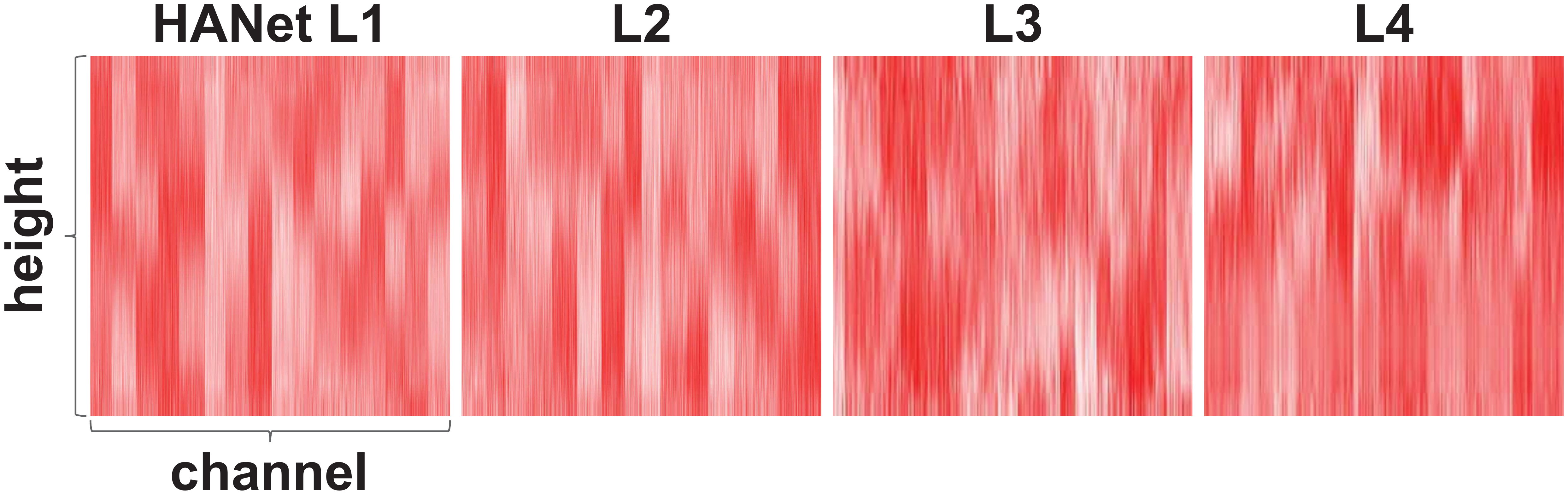}
\end{center}
\vspace*{-0.5cm}
   \caption{Visualization of attention map from HANet at different layers. x-axis denotes the channels, and y-axis denotes the height of the feature map, showing which channels are weighted at different vertical position. The channels focused by each height are clearly different. To better visualize, the channels are clustered.}
\label{fig:vis_attention}
\vspace*{-0.2cm}
\end{figure}

Fig.~\ref{fig:vis_attention} clearly shows that HANet assigns a different amount of attention to a different vertical position, indicating that the model properly learns structural priors with respect to the height in urban-scene data.

\begin{figure}[!b]
\vspace*{-0.4cm}
\begin{center}
  \includegraphics[width=0.99\linewidth]{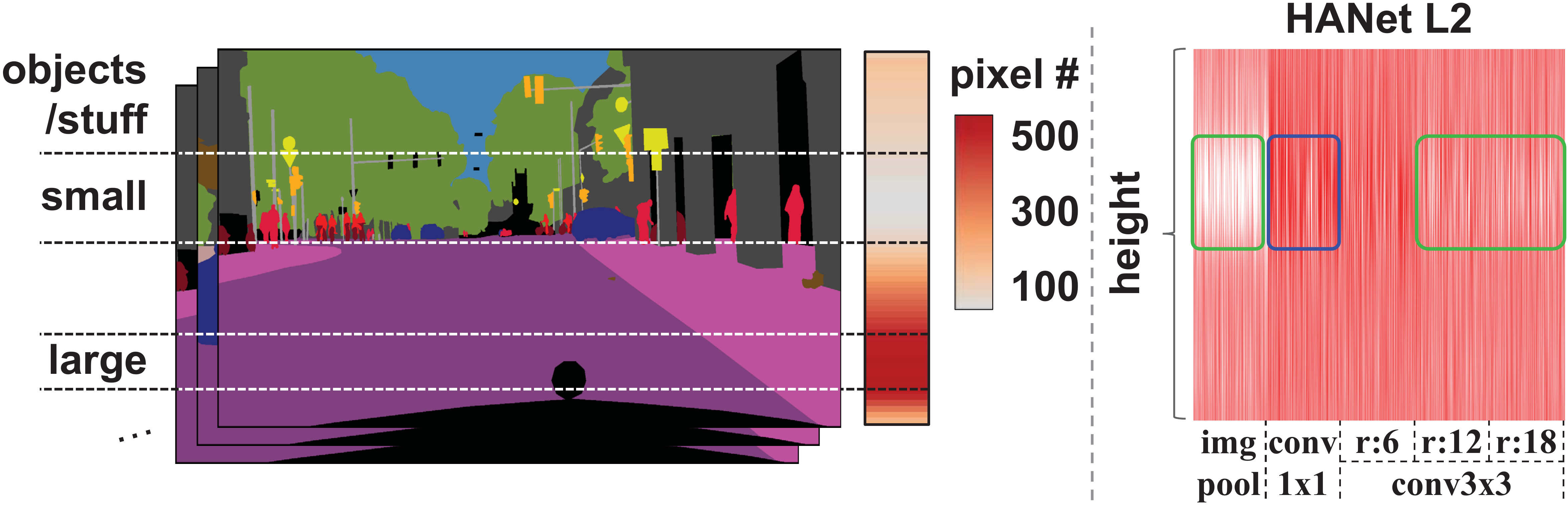}
\end{center}
\vspace*{-0.6cm}
   \caption{\textbf{(left)} The color in the heatmap denotes an average number of pixels that make up an object.
   Relatively small objects are crowded in the middle regions, while large objects exist in the lower region. 
   \textbf{(right)} Visualization of attention map from HANet at the second layer. Unlike Fig.~\ref{fig:vis_attention}, the sequence of channels remains unchanged.}
\label{fig:vis_aspp}
\end{figure}


Fig.~\ref{fig:vis_aspp} visualizes the attention map from HANet L2, which 
computes the attention weights for ASPP layer. In ASPP layer, the channels are obtained from convolution filters that have multi-scale receptive fields and grouped for each scale. Therefore, we can interpret the HANet L2 attention map by each group, with the sequence of channels remaining unchanged. Colored boxes give us insights of our method and 
the urban-scene images. The green boxes in Fig.~\ref{fig:vis_aspp} shows the low-focused channels of the middle region of images, while blue box indicates the channels which are relatively more focused. That is, the channels obtained from the small receptive field are weighted in the middle region. Since the middle region is where small objects are crowded as pointed in the
left figure in Fig.~\ref{fig:vis_aspp}, small receptive fields are effective for this region and vice versa. In this manner, we verify that HANet properly learns and captures the intrinsic features of urban scene.

\begin{figure}[h]
\vspace*{-0.1cm}
\begin{center}
  \includegraphics[width=0.85\linewidth]{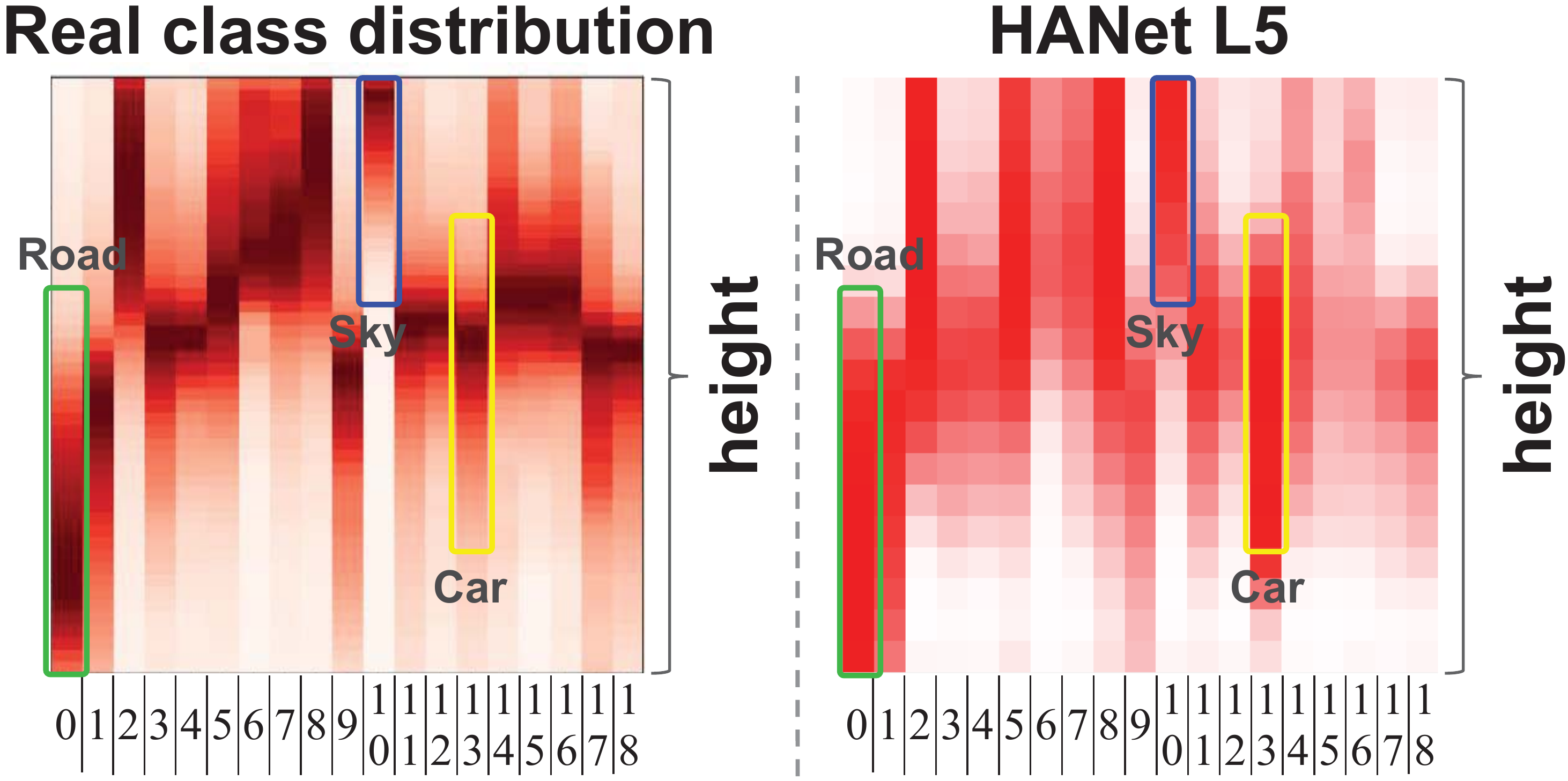}
\end{center}
\vspace*{-0.5cm}
   \caption{Height-wise class distributions and attention map visualization (L5). The number ranging from 0 to 18 indicates a different class. The darker it is colored, the higher probability (more pixels) assigned to a particular class.
    The attention visualization follows the patterns in the real class distribution.
   }
\label{fig:vis_last}
\vspace*{-0.5cm}
\end{figure}

Fig.~\ref{fig:vis_last} illustrates that the distribution of the attention map (right figure) from HANet at the last layer, which is following
the actual height-wise class distribution (left figure) obtained from the Cityscapes training images. Each class gives a different weight according to the vertical position,
meaning that the model actually uses vertical positions in the image. This information corresponds to
the observation we introduced through the class distribution analysis in Fig.~\ref{fig:intro} and Table~\ref{tab_intro}. For instance, road (class 0) appears
only in the middle and the lower regions, while sky (class 10) is mainly emphasized in the upper rows.

Visualization of semantic segmentation results are shown in the supplementary material.

\section{Conclusions}
\vspace*{-0.1cm}
In this paper, we proposed HANet for urban-scene segmentation as an effective and wide applicable add-on module. 
Our method exploits the spatial priors existing in urban-scene images
to construct a cost-effective architecture. We demonstrated the performance increase by adding our method to the baseline model with negligible cost increase.
Moreover, we visualized and analyzed the attention maps to show the validity of our initial hypothesis that exploiting vertical positional information helps for urban-scene semantic segmentation. 

\vspace*{0.2cm}
{
\noindent\textbf{Acknowledgments} 
This work was partially supported by Institute of Information \& communications Technology Planning \& Evaluation (IITP) grant funded by the Korea government (MSIT) (No.2019-0-00075, Artificial Intelligence Graduate School Program (KAIST)), the National Research Foundation of Korea (NRF) grant funded by the Korean government (MSIP) (No. NRF-2018M3E3A1057305), and the National Supercomputing Center with supercomputing resources including technical support (KSC-2019-INO-0011).  
}




\clearpage

{\small
\bibliographystyle{latex/ieee_fullname}
\bibliography{main}
}

\clearpage

\def\thesection{\Alph{section}}

\setcounter{section}{0}
\section{Supplementary Material} \label{supple}

This section complements our paper by presenting additional information, experimental results and visualizations. First, we provide further comparison results with other state-of-the-arts in Section~\ref{app:1_comparison}. In Section~\ref{app:position}, we conduct experiments to find the best way to incorporate positional information. We then describe the architecture details of the baseline and HANet in Section~\ref{app:further_details}. In Section~\ref{app:height_wdith}, we compare height-wise and width-wise class distributions. Finally, we conduct quantitative and qualitative comparisons between ours and the baseline model in Section~\ref{app:segmentation_maps}.

\subsection{Additional comparisons with other models} \label{app:1_comparison}
\vspace*{-0.1cm}
We compare the best performance of our model with other state-of-the-arts on the Cityscapes validation set.

\begin{table}[!h]
\vspace{-0.1cm}
\begin{center}
\footnotesize
\setlength\tabcolsep{10pt}
\begin{tabular}{c|c|c}
\toprule
Models (Year) & Backbone & mIoU(\%) \\
\drule
ANLNet~\cite{zhu2019asymmetric} ('19) & ResNet-101 & 79.9 \\
\midrule
DANet~\cite{fu2019dual} ('19) & ResNet-101 & 81.5 \\
\midrule
CCNet~\cite{huang2019ccnet} ('19) & ResNet-101 &  81.3 \\
\midrule
ACFNet~\cite{zhang2019acfnet} ('19) & ResNet-101 &  81.46 \\
\drule
Ours & ResNet-101 & \textbf{82.05} \\
\bottomrule
\end{tabular}
\end{center}
\vspace*{-0.5cm}
\caption{Comparisons against the best performances reported in the published papers of other state-of-the-art models on the Cityscapes validation set. The models based on ResNet-101 are compared.}
\label{tab_val_comparisons}
\vspace*{-0.4cm}
\end{table}


\subsection{Positional encoding and embedding.} \label{app:position}
\vspace*{-0.1cm}
In the NLP domains, there exist different approaches to inject positional information of each token in the input sequence. Positional encoding using sinusoidal values~\cite{vaswani2017attention} and learned positional embeddings~\cite{gehring2017convolutional} have been shown to produce comparable performances~\cite{vaswani2017attention}. We conduct experiments to find the best way to incorporate positional information. It turns out that the best way is to put sinusoidal positional encoding into the second convolutional layer of HANet (Table~\ref{tab_poisitional}).

\begin{table}[!h]
\vspace{-0.1cm}
\begin{center}
\footnotesize
\begin{tabular}{c|cc}
\toprule
\multirow{2.5}{*}{Methods} & \multicolumn{2}{c}{Injected layer}  \\\cmidrule{2-3}
& 1st & 2nd \\
\drule
Sinusoidal encoding & 79.61\% & \textbf{80.29}\% \\
\midrule
Learnable embedding (from scratch) & 79.95\% & 79.60\%  \\
\midrule
Learned embedding (from pretrained) & 79.61\% & 79.30\%  \\
\bottomrule
\end{tabular}
\end{center}
\vspace*{-0.5cm}
\caption{Performances comparison with respect to the layers and methods of positional encoding. Note that HANet consists of three convolutional layers. ResNet-101, output stride 8 on Cityscapes validation set.}
\label{tab_poisitional}
\vspace*{-0.4cm}
\end{table}

\begin{figure}[!t]
\centering
  \includegraphics[width=0.7\linewidth]{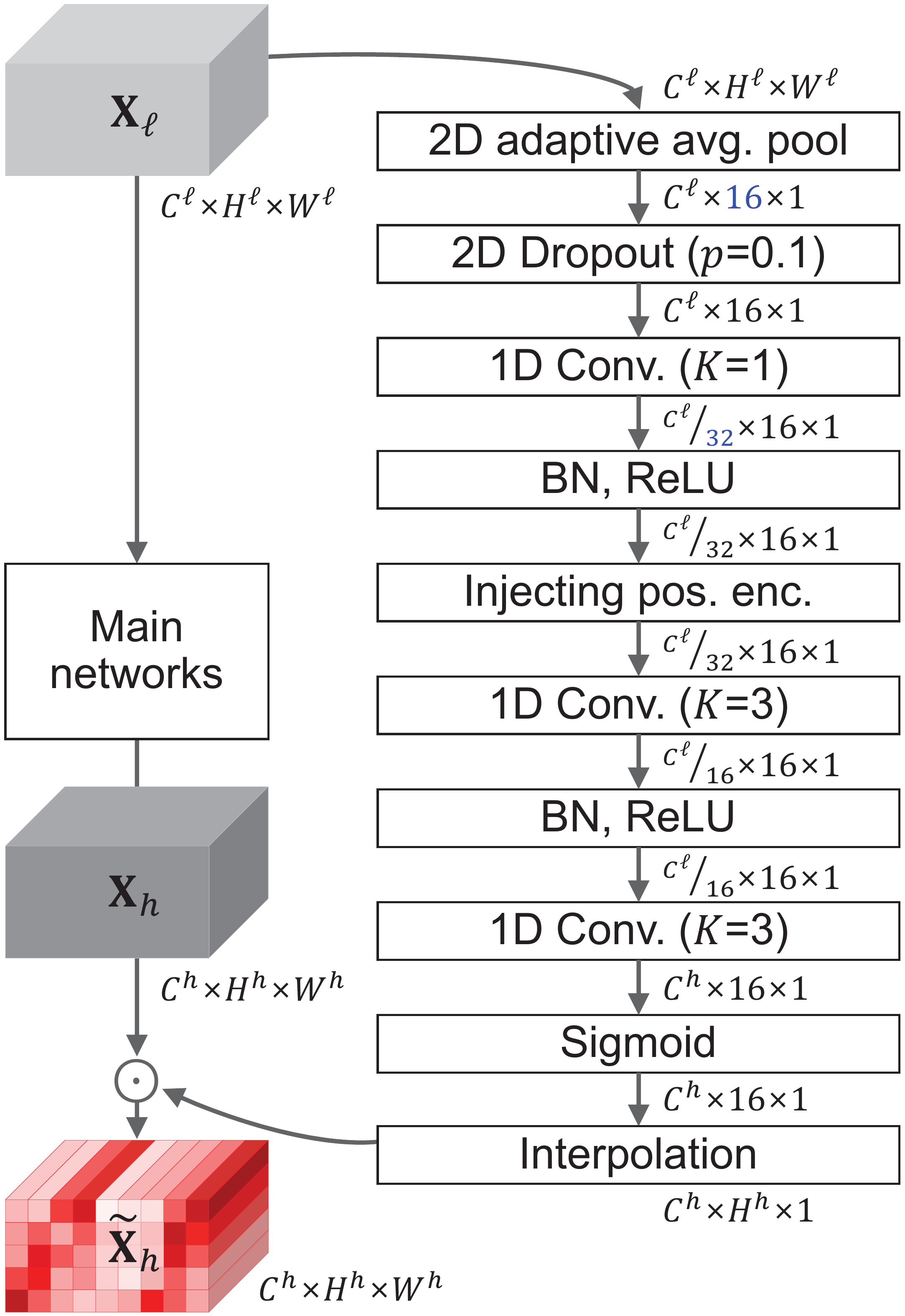}
  \vspace*{-0.1cm}
  \caption{Detailed architecture of HANet. $p$ denotes the dropout probability, and $K$ denotes the kernel size of each one-dimensional convolution layer. BN denotes a batch normalization layer.}
\label{fig:hanet_arch}
\vspace*{-0.5cm}
\end{figure}

\subsection{Further implementation details} \label{app:further_details}
\vspace*{-0.1cm}
We implement our methods based on the open-source implementations of NVIDIA semantic segmentation model~\cite{zhu2019improving}.
HANet consists of three convolutional layers incorporating dropout and batch normalization. To extract the height-wise contextual information from each row, we empirically adopt average pooling.
\vspace*{-0.3cm}
\paragraph{HANet architecture}
Fig.~\ref{fig:hanet_arch} shows detailed architecture of HANet. Width-wise pooling and interpolation for coarse attention are implemented using two-dimensional adaptive average pooling operation\footnote{\url{https://pytorch.org/docs/stable/nn.html\#torch.nn.AdaptiveMaxPool2d}} in Pytorch. 
Afterwards, a dropout layer and three one-dimensional convolutional layers are applied. Blue values in Fig.~\ref{fig:hanet_arch}, 16 and 32, are respectively the height of coarse attention and the channel reduction ratio $r$, which are our hyper-parameters. All the hyperparameters can be found in our code.

\begin{figure}[t]
\centering
  \includegraphics[width=1.03\linewidth]{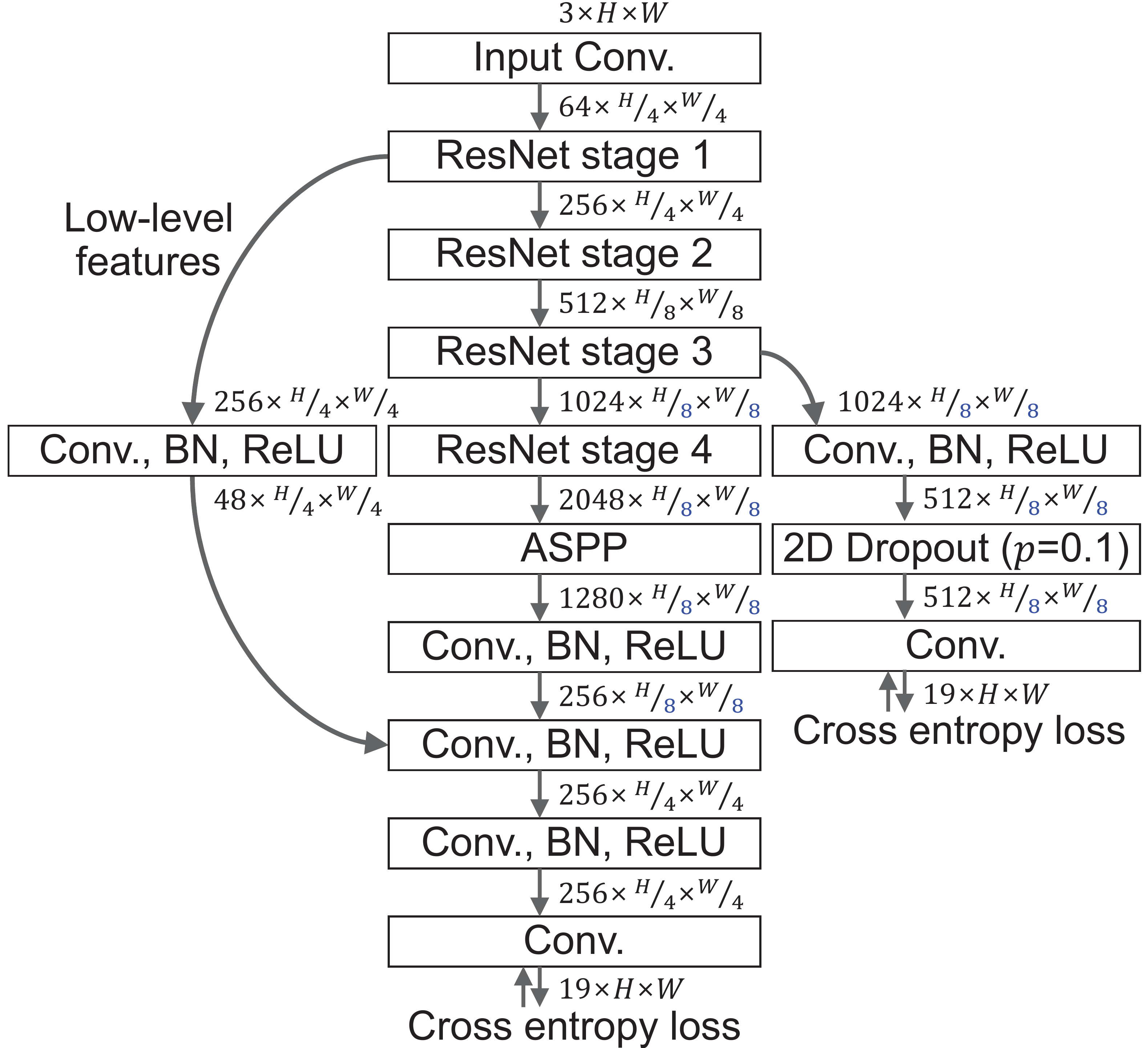}
  \vspace*{-0.6cm}
  \caption{Detailed architecture of the baseline model}
\label{fig:baseline_arch}
\vspace*{-0.3cm}
\end{figure}

\vspace*{-0.4cm}
\paragraph{Baseline architecture}
Fig.~\ref{fig:baseline_arch} shows detailed architecture of the baseline model, which is based on DeepLabv3+. As an encoder-decoder architecture, low-level features obtained from ResNet stage 1 are concatenated to high-level features via skip-connection. An auxiliary loss proposed in PSPNet~\cite{li2018pyramid} is applied to facilitate the learning process of deep networks. To adopt the auxiliary loss, additional convolutional layers are added after ResNet stage 3 as an auxiliary branch. The loss for this auxiliary branch has a weight of 0.4.
The output stride is set to 8 as shown in blue color; this can be set differently, e.g., 16.

\begin{table*}[!t]
\vspace*{-0.3cm}
\begin{center}
\setlength\tabcolsep{3.9pt}
\footnotesize
\begin{tabular}{c|c|ccccccccccccccccccc}
\toprule
Model & mIoU & road & swalk & build. & wall & fence & pole & tligh. & tsign & veg & terr. & sky & pers. & rider & car & truck & bus & train & mcyc & bcyc\\
\drule
Baseline & 81.14 & 98.5 & 87.3 & 93.6 & 66.1 & 64.4 & 68.7 & 74.0 & 82.0 & 93.2 & 65.6 & 95.2 & 84.3 & 66.0 & 95.7 & 80.6 & 92.8 & 85.0 & 68.9 & 80.0\\
$+$HANet & 82.05 & 98.6 & 87.7 & 93.7 & 66.7 & \textbf{66.2} & 68.7 & 74.4 & 81.9 & 93.3 & \textbf{67.7} & 95.3 & 84.5 & 66.9 & 96.1 & \textbf{87.9} & 92.7 & \textbf{86.0} & \textbf{70.7} & 80.1 \\
\bottomrule
\end{tabular}
\end{center}
\vspace*{-0.6cm}
\caption{Performance comparison of our methods against the baseline in terms of per-class IoU and mIoU measures. Inference techniques such as sliding, multi-scale, and flipping are applied. ResNet-101, output stride 8 on the Cityscapes validation set.
}
\label{tab_experiment}
\vspace*{-0.2cm}
\end{table*}

\subsection{Height- and width-wise class distribution} \label{app:height_wdith}
\begin{figure}[!t]
\centering
  \includegraphics[width=\linewidth]{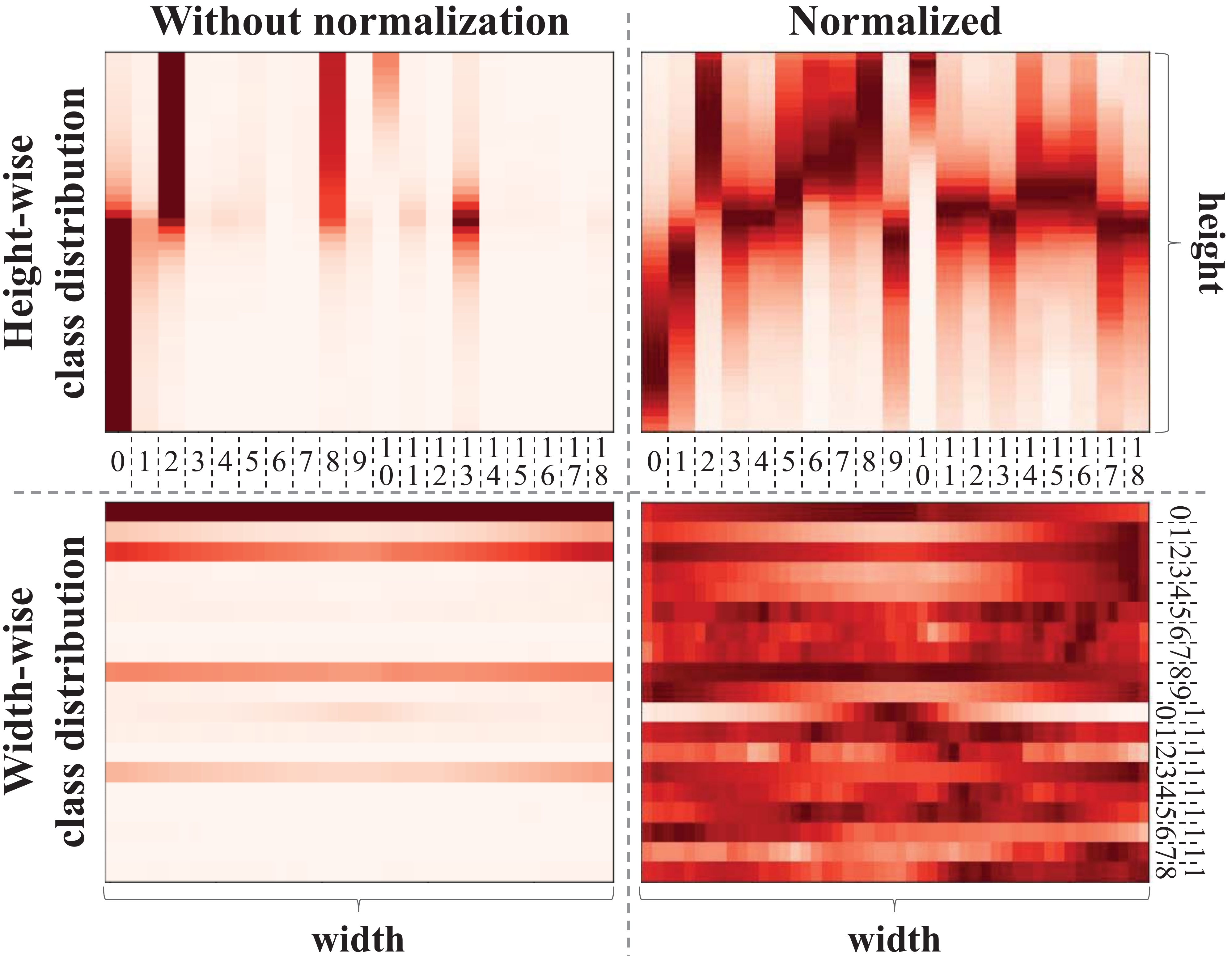}
  \caption{Comparison of a height-wise and a width-wise class distributions. 
  A darker color indicates a higher probability (more pixels) assigned to a particular class (from 0 to 18). The height-wise class distributions show distinct patterns across vertical positions while it is not the case for width-wise ones. Normalized distributions of each class are presented on the right column.
  }
\label{fig:row_vs_col}
\vspace*{-0.3cm}
\end{figure}

\vspace*{-0.1cm}
As shown in Fig.~\ref{fig:row_vs_col}, the width-wise class distributions are relatively similar across columns than the height-wise ones are, so it would be relatively difficult to extract distinct information with respect to the horizontal position of an image. 
Also, empirically, no meaningful performance increase has been observed when using the attention networks exploiting a width-wise class distribution.

This clear pattern corroborates the rationale behind the idea of HANet that extracts and incorporates height-wise contextual information rather than the width-wise one.

\subsection{Per-class IoU and segmentation maps} \label{app:segmentation_maps}
\begin{figure*}[!t]
\centering
  \includegraphics[width=\linewidth]{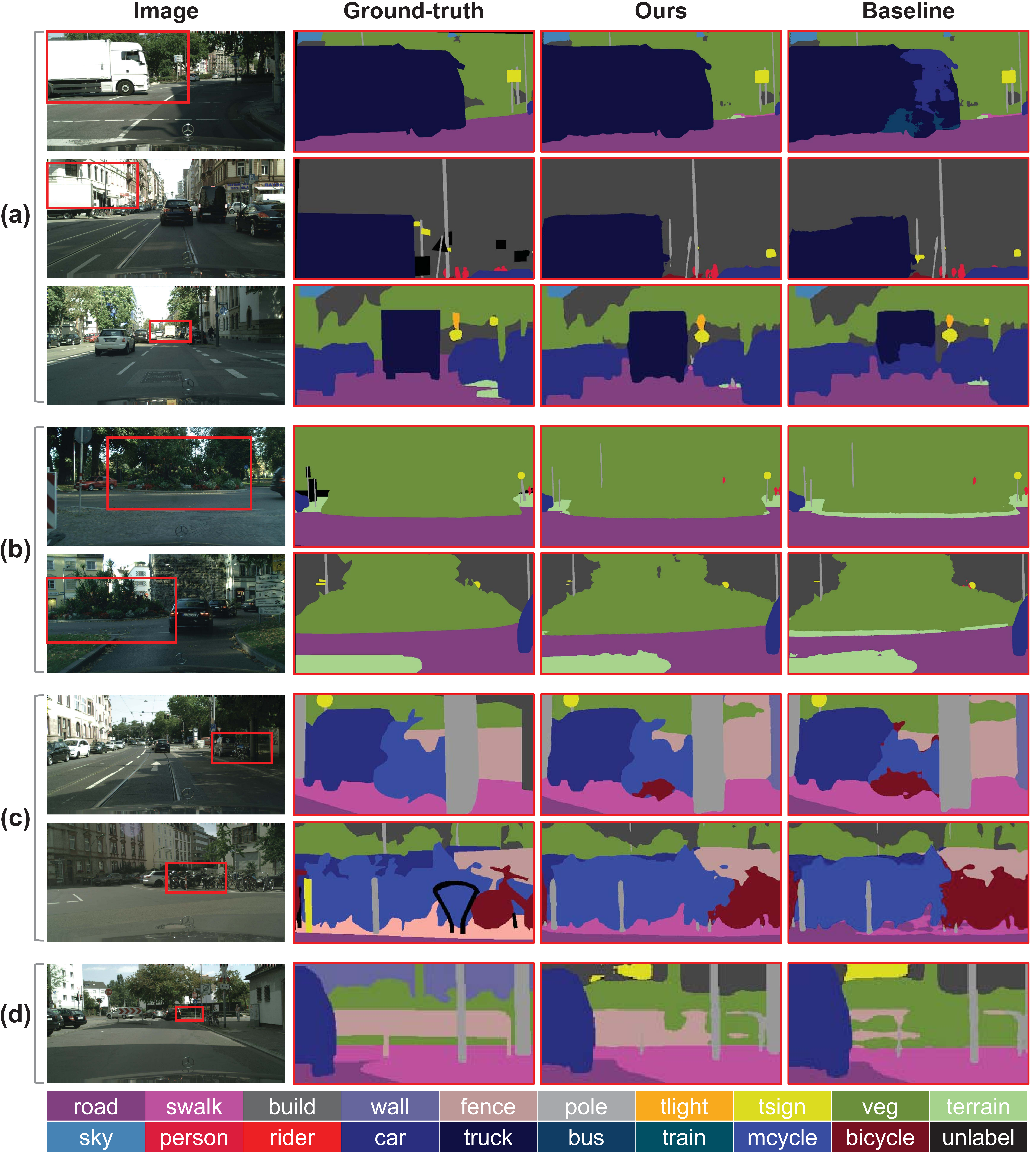}
  \caption{Comparison of predicted segmentation maps: (a) truck, bus, and car. (b) vegetation and terrain. (c) motorcycle and bicycle. (d) fence and vegetation.}
\label{fig:vis_result_1}
\vspace*{+1.0cm}
\end{figure*}

\begin{figure*}[!t]
\centering
  \includegraphics[width=\linewidth]{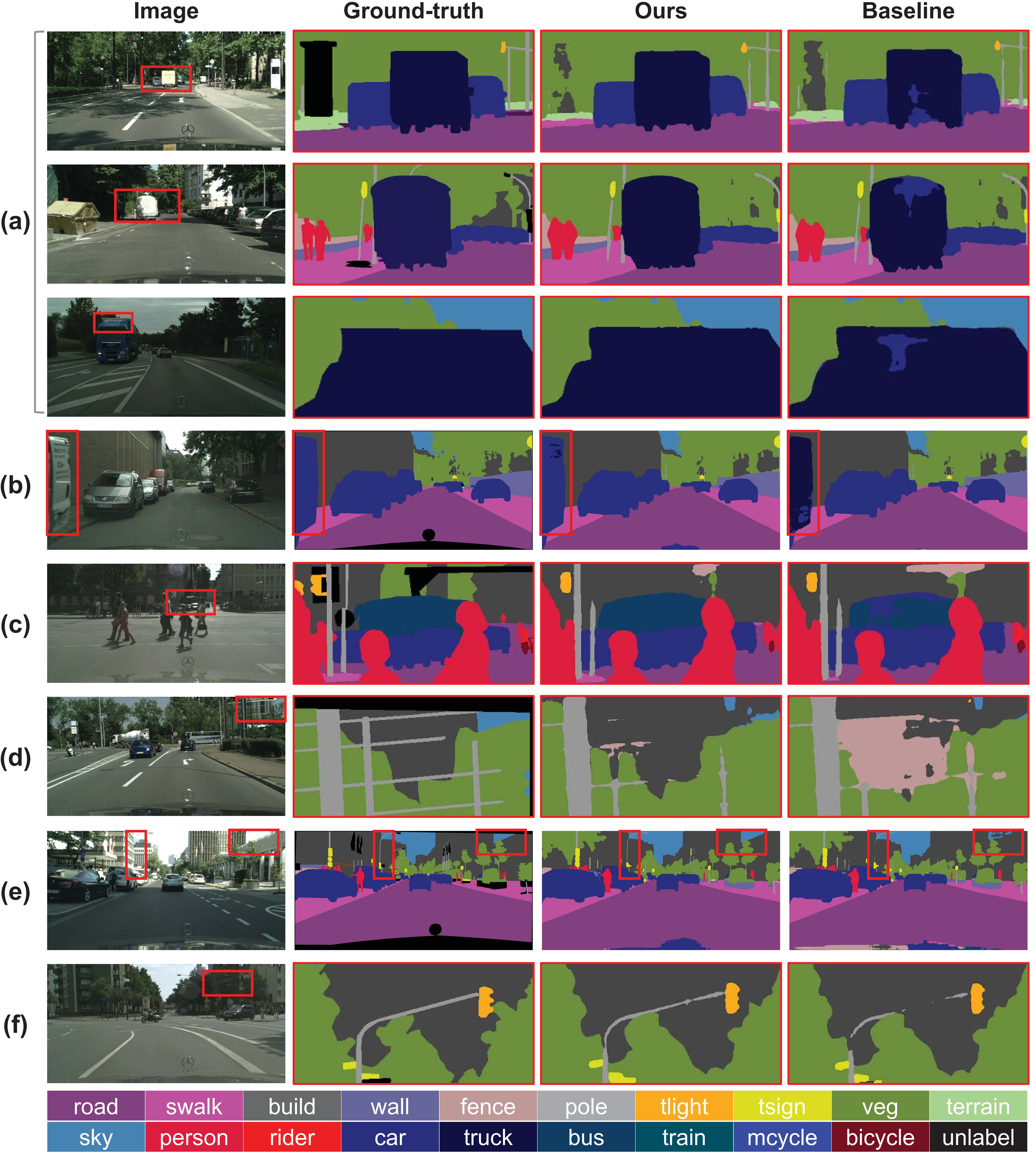}
  \caption{Comparison of predicted segmentation maps: (a) truck and car. (b) car and truck. (c) bus and car. (d) building and fence. (e) sky and building; pole and building. (f) pole and building.}
\label{fig:vis_result_2}
\vspace*{+1.0cm}
\end{figure*}

\vspace*{-0.1cm}
We present per-class IoU and segmentation maps to analyze HANet qualitatively and quantitatively. 
\vspace*{-0.4cm}
\paragraph{Comparison to baseline.}
Table~\ref{tab_experiment} shows the per-class IoU and mIoU results to compare the baseline and our methods in detail. 
Compared to the baseline, all the classes show similar or improved results; up to 7.3\% IoU increase is observed.
Qualitatively, ours can properly distinguish individual objects, even between the classes much alike to each other (Figs.~\ref{fig:vis_result_1} and \ref{fig:vis_result_2}). 
From our result in Fig.~\ref{fig:vis_result_1}(a) and Fig.~\ref{fig:vis_result_2}(a)-(c), one can see that the train, trucks, or cars in a far, crowded region are properly predicted by ours, even if similar vehicles are found nearby. 
Also, vegetation is accurately distinguished from terrain in ours, compared to the baseline (Fig.~\ref{fig:vis_result_1}(b)). 
Another interesting examples are found in  Fig.~\ref{fig:vis_result_2}(e)-(f);
the poles are connected fully in ours but dotted or missed in the baseline. 
We conjecture that HANet helps to distinguish confusing classes by properly gating the activation maps using height-wise contextual information based on their vertical positions. To summarize, compared to the baseline, our method generally forms a clear boundary of a object while avoiding its unnecessary fragmentation into multiple pieces.

\end{document}